\documentclass[runningheads]{llncs}

\usepackage[T1]{fontenc}
\usepackage{graphicx}
\usepackage{booktabs}
\usepackage{amssymb}
\usepackage{array}
\usepackage{xcolor}            
\usepackage{bbding}            
\usepackage{tikz}              
\usetikzlibrary{shapes.geometric, arrows.meta, positioning, fit, calc}
\usepackage{url}
\usepackage[hidelinks]{hyperref}          

\newcolumntype{L}[1]{>{\raggedright\arraybackslash}p{#1}}

\begin{document}

\title{PolicyKG: An Agentic LLM Pipeline for Translating Institutional
       Policies into SHACL Knowledge Graphs}
\titlerunning{PolicyKG}

\author{%
  Ponkrit Kaewsawee\textsuperscript{(\Envelope)}\inst{1}\orcidID{0009-0002-8382-4584} \and
  Chaklam Silpasuwanchai\inst{1}\orcidID{0000-0002-8670-6893} \and
  Chutiporn Anutariya\inst{1}\orcidID{0000-0001-7101-212X}}
\authorrunning{P. Kaewsawee et al.}

\institute{Asian Institute of Technology, Pathum Thani, Thailand\\
\email{ponkrit.ka@spu.ac.th}\\
\email{\{chaklam,chutiporn\}@ait.asia}}

\maketitle

\begin{abstract}
Institutional policies are written in natural language, but verifying
compliance increasingly demands executable constraints, leaving
organisations to bridge that gap by hand. We present \textsc{PolicyKG},
an agentic LLM pipeline that translates institutional policy documents
into SHACL-validated knowledge graphs via a first-order deontic
intermediate representation, orchestrated as four self-correcting stages
over a LangGraph state machine. The main technical contribution is the
\emph{Corpus Adapter}, a configuration-driven vocabulary-grounding layer
that constrains LLM-generated predicates to a target ontology and
enables cross-domain transfer by swapping registries rather than
retraining models. Evaluated on the full Asian Institute of Technology
Policies~\&~Procedures corpus (1{,}663 sentences yielding 443 rules),
PolicyKG achieves 86.9\% deontic-type classification accuracy
(Cohen's $\kappa$ = .709, corroborated by a three-annotator IRR with
Fleiss' $\kappa$ = .844, item-level bootstrap 95\% CI $[.67, .97]$)
and SHACL shape correctness of \textit{F}\textsubscript{1} = .866,
95\% CI $[.79, .93]$, on a 69-shape evaluation subset of an
ontology-grounded reference set. End-to-end formalisation yield is
79.2\% (351/443) via the FOL path; the remaining 20.8\% fall back to
a direct NL$\to$SHACL path whose failure modes are predominantly
structural (long compound conjunctions, embedded bullets), not
expressiveness-bounded. A two-pass audit (an automated regex checklist over all 443 rules
plus a first-author manual review of the 92 FOL-fallback cases where
higher-order constructs could otherwise hide) flagged zero
second- or higher-order constructs (exact Clopper--Pearson upper
95\%~CI on the true HOL requirement rate: 0.67\%). We report this
as an audit finding for this corpus, not a proof of FOL sufficiency
in general. A controlled cross-domain experiment on
\textit{N} = 15 GDPR rules raises exact property alignment from 1/15
to 11/15 with the Corpus Adapter (Fisher's exact \textit{p} < .001;
Cohen's \textit{h} = 1.53, large effect). External validation on the
LexDeMod~\cite{sancheti2022lexdemod} lease-contract benchmark
(\textit{N} = 200) drops Macro \textit{F}\textsubscript{1} from .810
to .370, with permission \textit{F}\textsubscript{1} = .038; the
failure mechanism is a vocabulary-prior mismatch (lease English uses
\emph{shall be entitled} for permission) precisely of the class the
Corpus Adapter is designed to repair by registry swap. We release
the gold standard, both vocabulary registries, evaluation scripts,
and pipeline configuration; repeated runs on the same host produce
hash-identical SHACL outputs.

\end{abstract}

\section{Introduction}
\label{sec:intro}
Institutional policies, which govern students, staff, and operations in
universities, hospitals, and government agencies, are written as
natural-language documents and read by humans. Yet the work of
\emph{verifying} that institutional practice conforms to those policies is
increasingly automated: enrolment workflows, payment systems, and
accreditation reports all consume machine-readable data. The gap between
human-authored policy and machine-checked compliance is bridged today by
manual interpretation: a registrar reads a paragraph, decides what it
means, and configures a system rule. This process does not scale, does not
survive policy revisions, and offers no audit trail linking a system check
back to the textual obligation it enforces.

Three lines of research speak to this problem. Policy and legal NLP have
produced rule-extraction systems based on patterns~\cite{maxwell2012},
classical classifiers~\cite{brodie2006}, and recently large language
models~\cite{goknil2024}, reporting accuracies between 75\% and 85\% on
isolated extraction tasks. Deontic logic, developed over more than half a
century, offers operators of obligation, permission, and prohibition that
capture the normative content of policy text~\cite{governatori2010,mcnamara2006}.
SHACL, the W3C Shapes Constraint Language~\cite{w3c2017}, encodes
validation constraints over RDF knowledge graphs and is increasingly used
for compliance checking~\cite{palmirani2018}. Each of these threads has
matured in isolation. What is missing is an empirically validated,
end-to-end pipeline that connects natural-language policy to executable
SHACL, together with a satisfying answer to the open theoretical
question of whether first-order logic with deontic extensions is \emph{sufficient} for
institutional policy, or whether higher-order constructs are required.

We present \textbf{PolicyKG}, an agentic LLM pipeline that translates
institutional policy documents into SHACL-validated knowledge graphs. The
pipeline runs four orchestrated stages: document extraction, deontic rule
classification, FOL${+}$deontic formalisation, and SHACL translation.
A LangGraph state machine coordinates them, with per-stage validators
that trigger self-correction on local failure.
We also introduce the \textbf{Corpus Adapter}: a configuration-driven
vocabulary-grounding layer that constrains LLM generation to a target
ontology and enables the same pipeline to retarget across domains by
swapping the registry. We evaluate PolicyKG on the full Asian Institute of
Technology (AIT) Policies \& Procedures corpus, which contains 1{,}663
sentences yielding 443 deontic rules, and demonstrate cross-domain transfer to a GDPR-derived
corpus.

\paragraph{Contributions.}
\begin{itemize}
  \item An end-to-end agentic LLM pipeline
        (NL${\to}$FOL${+}$deontic${\to}$SHACL) validated on a real
        1{,}663-sentence institutional corpus.
  \item The \emph{Corpus Adapter}, a vocabulary-grounding pattern that
        enables cross-domain transfer. A controlled GDPR experiment
        (\textit{N} = 15) raises exact property alignment from 1/15 to 11/15
        (Fisher's exact \textit{p} < .001; Cohen's \textit{h} = 1.53).
  \item A two-pass empirical audit finding that first-order deontic
        logic is observationally sufficient on this corpus: an
        automated regex checklist flags 0 of 443 rules with
        second- or higher-order construct
        patterns~\cite{governatori2010}, and a first-author manual
        review of the 92 FOL-fallback rules (where HOL requirements
        could otherwise hide behind LLM parse failure) confirms
        0/92 genuine HOL cases (exact Clopper--Pearson upper 95\%~CI
        on the true HOL rate: 0.67\%). Framed as an audit finding
        rather than a proof.
  \item An open evaluation: three-annotator gold (Fleiss'
        $\kappa$ = .844, item-level bootstrap 95\% CI $[.67, .97]$),
        LexDeMod external validation, and fixed-seed reproducibility
        yielding hash-identical SHACL outputs.
\end{itemize}

\section{Related Work}
\label{sec:related}

\subsection{Policy and Legal NLP}
Pattern-based extraction was the first generation of automated policy NLP.
Maxwell and Ant\'on~\cite{maxwell2012} applied regular expressions and
keyword templates to privacy policies, reporting around 77\% accuracy but
struggling with syntactic variation. Brodie et al.~\cite{brodie2006}
improved on this using SVM classifiers over hand-crafted features,
reaching 82\% on financial regulation but at a feature-engineering cost
that did not transfer across document types. Subsequent hybrids of pattern matching and ontology grounding reached
~85\% accuracy on legal text but at the price of substantial upfront
ontology development. The common limitation is reliance on bespoke
features or hand-built ontologies; none draws on the semantic priors
of pre-trained language models.

The recent LLM era has produced benchmarks but few end-to-end systems.
LexGLUE~\cite{chalkidis2022} established legal-language-understanding
baselines across multiple tasks; LegalBench~\cite{guha2023} expanded to
162 legal-reasoning tasks. PAPEL~\cite{goknil2024} applied
chain-of-thought prompting to privacy-policy annotation, reaching 80\%
F1 on clause labelling. These benchmarks define the landscape but do not,
individually, produce executable SHACL constraints from policy text.

\subsection{Deontic Logic for Normative Systems}
Deontic logic provides a mathematically rigorous account of obligation,
permission, and prohibition~\cite{mcnamara2006}. Governatori and
Rotolo~\cite{governatori2010} argued, on theoretical grounds, that
first-order deontic logic supplies sufficient expressiveness for
normative systems in legal and institutional contexts. The open question, whether real institutional policies actually stay
within FOL${+}$deontic or reach into SOL/HOL territory, was framed
by the standard articulation of the expressiveness/tractability trade-off~\cite{brachman2004}.
Until the present work this question had been addressed only at the
theoretical level; we provide the first empirical audit finding on a
real institutional corpus (\S\ref{sec:results-rq2}), scoped to the
tested sub-domain and framed as an observational lower bound on FOL
adequacy rather than a theoretical resolution.
Standard Deontic Logic is known to be subject to the Chisholm and Ross
paradoxes, and to handle conditional exceptions
(contrary-to-duty obligations, waiver clauses) only awkwardly. The
established response is defeasible deontic
logic~\cite{nute1994defeasible,governatori2018ddl}, with argumentation-
theoretic accounts of legal reasoning~\cite{prakken2015defeasible}
providing the broader framework. The interchange format
\textbf{LegalRuleML}~\cite{athan2013legalruleml} standardises the
serialisation of such rules with defeasibility, override, and
temporal-scope metadata. PolicyKG does not attempt a full defeasible
account; it records override relations via a custom
\texttt{deontic:overrides} annotation (\S\ref{sec:method}) as a
stepping stone that a downstream defeasible layer could consume, and
leaves full LegalRuleML alignment to future work
(\S\ref{sec:conclusion}).

\subsection{SHACL for Compliance}
The W3C Shapes Constraint Language~\cite{w3c2017} enables declarative
validation of RDF knowledge graphs through node and property shapes
annotated with severity levels. Palmirani and
Governatori~\cite{palmirani2018} argued that SHACL's first-order pattern
language has adequate expressiveness for institutional policy
formalisation, encoding obligations, permissions, and prohibitions via
severity and custom annotations. Their argument was theoretical and
lacked empirical validation, and no prior pipeline produces SHACL
automatically from natural-language policy text at the scale of an
institutional corpus. PolicyKG provides the empirical realisation their
work anticipated.

\subsection{LLMs for Structured Output and Knowledge Graph Construction}
A growing body of work uses LLMs to construct or query knowledge
graphs.
Schema-constrained and grammar-guided
generation~\cite{willard2023outlines,geng2023grammar} forces LLM
outputs into well-typed structures, addressing the \emph{syntactic}
side of structured output. Chain-of-thought
prompting~\cite{wei2022cot} is the dominant technique for
multi-step reasoning tasks and underpins Stage~3's FOL
formalisation prompt in this paper.
Retrieval-augmented generation (RAG)~\cite{lewis2020rag} is the
mainstream approach for grounding LLM outputs in an external
knowledge source at inference time, and
GraphRAG~\cite{graphrag2024} extends the pattern to knowledge-graph
retrieval, but it assumes the target graph already exists.
The \emph{semantic} side of \emph{constructing} that graph, that is,
ensuring LLM-generated predicates align with a target ontology rather
than hallucinated English property names, receives less direct
treatment in the literature. The Corpus Adapter
(\S\ref{sec:adapter}) is our answer to this semantic gap:
constrained selection from a configuration artefact in place of
either fine-tuning or open generation.
\S\ref{sec:results-gdpr} reports a head-to-head comparison against
top-\textit{K} RAG on the ontology property list, showing that RAG
closes most of the vocabulary-alignment gap; the Corpus Adapter's
remaining advantage is in \emph{deployment properties} (no embedding
dependency, deterministic, config-swappable) rather than raw
accuracy.

\subsection{Positioning}
The gap PolicyKG addresses sits at the intersection of these four
threads: an empirically validated, end-to-end pipeline that combines
(a)~LLM-based rule identification and formalisation, (b)~FOL${+}$deontic
semantics with an empirical audit-finding of sufficiency on a real
institutional corpus (a lower bound, not a proof), (c)~SHACL as the
operational compliance target, and (d)~a portable vocabulary-grounding
mechanism that mediates between LLM-natural predicates and target
ontologies. No prior system unifies these four components, and no
prior empirical work has audited the FOL-sufficiency question on a
real corpus.

\section{Preliminaries}
\label{sec:prelim}

\subsection{First-Order Deontic Logic}
We work in first-order logic extended with three deontic modal operators.
The syntax assumes constants, predicates of arity one or more, the
quantifiers $\forall$ and $\exists$, the connectives $\land, \lor,
\rightarrow, \neg$, and three modal operators applied to closed formulae:
$O(\varphi)$ asserts that $\varphi$ is obligatory; $P(\varphi)$ that
$\varphi$ is permitted; and $F(\varphi)$ that $\varphi$ is forbidden, with
$F(\varphi) \equiv O(\neg \varphi)$. We adopt the Standard Deontic Logic
axioms~\cite{mcnamara2006}, in particular $O(\varphi) \rightarrow
P(\varphi)$, which captures the intuition that what is required is
implicitly allowed. As a worked example, the policy sentence
\emph{``Students must register before classes begin''} admits the
formalisation
\begin{equation}
  \forall x\,\bigl( \mathit{Student}(x) \,\rightarrow\,
    O(\mathit{RegisterBefore}(x, \mathit{Classes}))\bigr).
  \label{eq:reg-obligation}
\end{equation}
\noindent
Equation~\ref{eq:reg-obligation} is the canonical pattern for a
universally quantified obligation; permissions replace $O$ with $P$ and
prohibitions replace $O$ with $F$.

\subsection{SHACL in One Page}
SHACL constraints validate RDF~\cite{w3c2014} graphs. A \texttt{NodeShape}
targets a class
via \texttt{sh:targetClass} and bundles a list of property shapes; a
\texttt{PropertyShape} constrains a specific predicate path
(\texttt{sh:path}) with cardinality (\texttt{sh:minCount},
\texttt{sh:maxCount}), value-type, or pattern restrictions. Violations are
reported at a configurable severity (\texttt{sh:Violation},
\texttt{sh:Warning}, or \texttt{sh:Info}). Equation~\ref{eq:reg-obligation}
translates to the SHACL fragment shown below:
\begin{verbatim}
ex:StudentRegShape a sh:NodeShape ;
  sh:targetClass ex:Student ;
  deontic:type deontic:obligation ;
  sh:severity sh:Violation ;
  rdfs:comment "Students must register before classes begin." ;
  sh:property [
    sh:path ex:registerBefore ;
    sh:minCount 1
  ] .
\end{verbatim}
\noindent
The custom \texttt{deontic:type} annotation preserves modal information
that would otherwise be lost in pure SHACL, and \texttt{rdfs:comment}
carries the original NL rule for audit. The \texttt{sh:Violation}
severity makes a missing \texttt{ex:registerBefore} link surface as a
compliance failure; a permission would instead emit an Info-level shape,
and a prohibition uses \texttt{sh:maxCount}\,0 on the forbidden property.

\subsection{The Translation Gap}
A naive approach, prompting an LLM to emit SHACL directly from policy
text, fails systematically along three axes. First, \emph{vocabulary
hallucination}: the LLM invents property names that do not exist in the
institutional ontology, so the generated shapes do not target any real
data. Second, \emph{lost deontic semantics}: the distinction between
obligation, permission, and prohibition collapses into uniform constraint
phrasing when severity assignment is left implicit. Third,
\emph{predicate-quality drift}: across re-runs the LLM produces
structurally similar but lexically divergent predicates, defeating
reproducibility. These three failure modes motivate the two main design
choices of PolicyKG: (i) an FOL${+}$deontic intermediate layer that makes
modality explicit and reviewable, and (ii) the Corpus Adapter
(\S\ref{sec:adapter}) that grounds vocabulary in the target ontology.

\section{The PolicyKG Pipeline}
\label{sec:method}

\subsection{Architecture Overview}
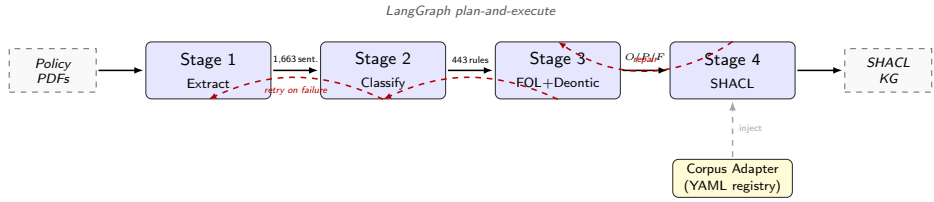
\begin{figure}[t]
  \centering
  \resizebox{\textwidth}{!}{%
  \begin{tikzpicture}[
      node distance=1.2cm and 0.9cm,
      stage/.style={rectangle, draw=black!85, rounded corners=3pt,
        minimum width=2.3cm, minimum height=1.05cm,
        align=center, font=\footnotesize\sffamily, fill=blue!10,
        inner sep=3pt, line width=0.4pt},
      io/.style={rectangle, draw=black!60, dashed,
        minimum width=1.6cm, minimum height=0.8cm,
        align=center, font=\scriptsize\sffamily\itshape,
        fill=gray!5, inner sep=2pt},
      adapter/.style={rectangle, draw=black!85, rounded corners=3pt,
        minimum width=2.2cm, minimum height=0.7cm,
        align=center, font=\scriptsize\sffamily,
        fill=yellow!20, inner sep=2pt, line width=0.4pt},
      arr/.style={-{Latex[length=1.7mm]}, thick, shorten >=1pt, shorten <=1pt},
      loop/.style={-{Latex[length=1.5mm]}, thick, dashed, red!70!black},
      lbl/.style={font=\tiny\sffamily, above, yshift=0.5pt}
    ]
    \node[io] (pdf) {Policy\\PDFs};
    \node[stage, right=of pdf] (s1) {Stage 1\\[1pt]\scriptsize Extract};
    \node[stage, right=of s1] (s2) {Stage 2\\[1pt]\scriptsize Classify};
    \node[stage, right=of s2] (s3) {Stage 3\\[1pt]\scriptsize FOL$+$Deontic};
    \node[stage, right=of s3] (s4) {Stage 4\\[1pt]\scriptsize SHACL};
    \node[io, right=of s4] (kg) {SHACL\\KG};

    \node[adapter, below=1.1cm of s4] (adapter) {Corpus Adapter\\[1pt]\scriptsize (YAML registry)};

    \draw[arr] (pdf) -- (s1);
    \draw[arr] (s1) -- node[lbl]{1{,}663\,sent.} (s2);
    \draw[arr] (s2) -- node[lbl]{443\,rules} (s3);
    \draw[arr] (s3) -- node[lbl]{$O$/$P$/$F$} (s4);
    \draw[arr] (s4) -- (kg);
    \draw[arr, dashed, gray!70] (adapter) -- (s4)
      node[midway, right, font=\tiny\sffamily]{inject};

    \draw[loop] (s2.south) to[bend right=25]
      node[below, font=\tiny\sffamily\itshape, yshift=-1pt]{retry on failure}
      (s1.south);
    \draw[loop] (s3.south) to[bend right=25] (s2.south);
    \draw[loop] (s4.north) to[bend left=35]
      node[above, font=\tiny\sffamily\itshape]{repair}
      (s3.north);

    \node[font=\scriptsize\sffamily\itshape, gray!70!black]
      at ($(s1.north)!0.5!(s4.north)+(0,0.55)$)
      {LangGraph plan-and-execute};
  \end{tikzpicture}%
  }
  \caption{PolicyKG pipeline. Four sequential stages
           (Stage~1--4, blue) are orchestrated by a LangGraph
           plan-and-execute layer that fixes the stage sequence for
           auditability; dashed red arrows show ReAct retry loops
           where per-stage validators route failed items back
           upstream with a corrective sub-prompt. The Corpus Adapter
           (yellow) is a YAML registry injected at Stage~4 only;
           retargeting to a new domain (\S\ref{sec:results-gdpr})
           swaps this registry with no other pipeline change.}
  \label{fig:pipeline}
\end{figure}

PolicyKG is structured as four sequential stages
(Fig.~\ref{fig:pipeline}): Stage~1 normalises a policy PDF into
sentences; Stage~2 classifies each sentence as one of \{Obligation,
Permission, Prohibition, Non-rule\}; Stage~3 lifts each classified rule
to an FOL${+}$deontic formula with top-level operator $\in\{O, P, F\}$;
Stage~4 maps each formula to a SHACL shape whose severity reflects
deontic modality. The four stages are orchestrated as a hybrid
\emph{ReAct}~+~\emph{Plan-and-Execute} agentic architecture on a
LangGraph state machine: plan-and-execute fixes the stage sequence for
auditability; ReAct loops inside each stage handle retry, with
per-stage validators routing failed items back upstream with a
corrective sub-prompt.

\subsection{Stage 1: Document Processing}
Policy PDFs are extracted with \texttt{pdfplumber} (headers, footers,
and page numbers stripped) and segmented by a policy-aware regex
splitter that handles list markers, soft line-wraps, and
semicolon-introduced clauses common in institutional P\&Ps. A
conservative simplification pass removes spurious mid-sentence line
breaks and normalises whitespace; modal verbs, quantifiers, and word
order are never altered. Semantic preservation is dual-checked by an
LLM self-assessment field and Sentence-BERT cosine similarity (threshold
$\geq 0.85$ accepts 97.9\% of rules). A one-to-one trace from each
simplified sentence to its source PDF span is preserved for audit. From
the five AIT P\&P documents (81 pages), Stage~1 yields 1{,}663
normalised sentences.

\subsection{Stages 2 and 3: Classification and FOL${+}$Deontic Formalisation}
Stage~2 uses a zero-shot prompt listing the four target labels with
short prototype examples and asks the LLM for a single label and a
confidence score on $[0,1]$, identical across all evaluated models. The
classifier is deliberately liberal; a downstream check in Stage~3
rejects any admitted sentence that fails to yield a well-formed deontic
formula. Stage~3 lifts each classified sentence into a first-order
formula with exactly one top-level operator from $\{O, P, F\}$ and the
modal-cue mappings (\textit{must, shall} ${\to}\,O$;\, \textit{may}
${\to}\,P$;\, \textit{must not, shall not} ${\to}\,F$); the prompt embeds
three demonstrations and constrains generation to a structured JSON
schema, addressing the constrained-decoding requirement that has
emerged as a standard remedy for LLM hallucination.
We deliberately do \emph{not} pass the Corpus Adapter vocabulary at
this stage; it is confined to Stage~4 so that Stage~3 produces a
representation whose predicate names are LLM-natural and human-readable,
decoupling deontic structure from vocabulary alignment. Each formula is
validated for well-formedness; a rejected formula is retried once with
the validator's diagnostic appended.

\begin{figure}[t]
  \centering
  \footnotesize\sffamily
  \begin{tabular}{@{}l@{\hspace{0.8em}}L{3.5cm}@{\hspace{0.8em}}L{3.5cm}@{\hspace{0.8em}}L{3.7cm}@{}}
    \toprule
     & \textbf{NL rule} & \textbf{FOL$+$deontic} & \textbf{SHACL shape} \\
    \midrule
    \textbf{O} &
      All students must pay tuition fees before each semester. &
      $\forall x.\, \mathrm{Student}(x) \to O(\mathrm{payFee}(x))$ &
      NodeShape($\mathrm{Student}$) + Property($\mathrm{payFee}$, $\mathrm{minCount}{=}1$, $\mathrm{sh{:}Violation}$) \\
    \textbf{P} &
      Students may reside off-campus from the second semester. &
      $\forall x.\, \mathrm{Student}(x) \to P(\mathrm{offCampus}(x))$ &
      NodeShape($\mathrm{Student}$) + Property($\mathrm{offCampus}$, $\mathrm{sh{:}Info}$) \\
    \textbf{F} &
      Students must not deface or vandalise library books. &
      $\forall x.\, \mathrm{Student}(x) \to F(\mathrm{deface}(x))$ &
      NodeShape($\mathrm{Student}$) + Property($\mathrm{deface}$, $\mathrm{maxCount}{=}0$, $\mathrm{sh{:}Violation}$) \\
    \bottomrule
  \end{tabular}
  \caption{NL${\to}$FOL${\to}$SHACL trace for the three deontic
           operators. Stage~4's deterministic mapping fixes the SHACL
           severity and cardinality.}
  \label{fig:trace}
\end{figure}

\subsection{Stage 4: SHACL Translation}
Translation from FOL to SHACL is rule-driven rather than LLM-driven,
because the mapping is structurally regular and we want auditable outputs.
The universal-obligation pattern
$\forall x\,(C(x) \to O(P(x,v)))$ becomes a NodeShape for $C$ and a
PropertyShape for $P$, with \texttt{sh:minCount}\,1 and severity
\texttt{sh:Violation}. Permissions $P(\varphi)$ become Info-level shapes
that surface allowed-but-not-required patterns. Prohibitions $F(\varphi)$
use \texttt{sh:maxCount}\,0 with severity \texttt{sh:Violation}, while
existential obligations use \texttt{sh:qualifiedValueShape}. Predicates
emitted by Stage~3 are
resolved against the Corpus Adapter's vocabulary registry at this
point (this is where vocabulary grounding takes effect), and any
unresolved predicate triggers a single repair pass through Stage~3 with
the adapter's constrained vocabulary fragment in-prompt.

Institutional policy contains conditional exceptions
(``the Registrar may waive the prerequisite in cases of documented
hardship'') that Standard Deontic Logic~\cite{mcnamara2006} handles
only awkwardly; the Chisholm and Ross paradoxes are the canonical
symptoms. Stage~4 records these as an explicit
\texttt{deontic:overrides} relation between shapes rather than
flattening the exception into a boolean predicate; the resulting
\texttt{ait:Rule\_A~deontic:overrides~ait:Rule\_B} triples are
first-class shape metadata that a downstream defeasible-reasoning
layer~\cite{governatori2018ddl} can consume. In the AIT corpus 12
override relations are emitted (\S\ref{sec:results-rq3}); a full
defeasible extension is deferred to future work
(\S\ref{sec:conclusion}).

\subsection{The Corpus Adapter}
\label{sec:adapter}
\begin{figure}[t]
  \centering
  \resizebox{\textwidth}{!}{%
  \begin{tikzpicture}[
      box/.style={rectangle, draw=black, rounded corners=2pt,
                  minimum width=2.3cm, minimum height=0.7cm,
                  font=\scriptsize\sffamily, align=center, inner sep=3pt},
      reg/.style={rectangle, draw=black!70, dashed,
                  minimum width=2.3cm, minimum height=0.7cm,
                  font=\scriptsize\sffamily, align=center, fill=yellow!10},
      arr/.style={-{Latex[length=1.3mm]}, thick},
      farr/.style={-{Latex[length=1.3mm]}, thick, dashed, gray!70},
      node distance=0.5cm and 0.5cm
    ]
    \node[reg] (reg) {Vocabulary registry\\ \scriptsize\itshape (YAML: classes, properties, synonyms)};

    \node[box, below=of reg] (fol) {FOL predicates\\ (Stage 3 output)};
    \node[box, right=2.0cm of fol, fill=blue!10] (sel) {Constrained selection\\ prompt to LLM};
    \node[box, right=of sel] (canon) {Post-hoc\\ canonicaliser};
    \node[box, right=of canon, fill=green!15] (shacl) {SHACL paths\\ aligned to ontology};

    \draw[farr] (reg.south) -- ++(0,-0.3) -| ([yshift=0.2cm]sel.north) node[midway, above, font=\tiny\itshape] {inject};
    \draw[farr] (reg.east) -- ++(2.0,0) -| ([yshift=0.2cm]canon.north);

    \draw[arr] (fol) -- (sel);
    \draw[arr] (sel) -- (canon);
    \draw[arr] (canon) -- (shacl);

    \draw[farr] (canon.south) -- ++(0,-0.4) -| (sel.south)
      node[midway, below, font=\tiny\itshape] {reject $\rightarrow$ Stage 3 repair};
  \end{tikzpicture}%
  }
  \caption{The Corpus Adapter is a thin three-part layer between
           Stage~3 (FOL formalisation) and Stage~4's SHACL emission:
           a YAML vocabulary registry, a constrained-selection prompt
           that converts open predicate generation into closed-set
           selection, and a post-hoc canonicaliser with a reject path
           that triggers a Stage~3 repair. Retargeting to a new domain
           swaps only the registry.}
  \label{fig:adapter}
\end{figure}
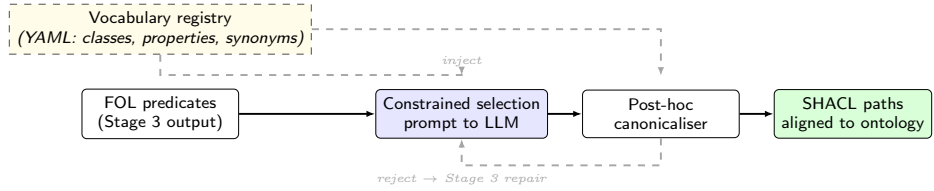

The Corpus Adapter is a thin, configuration-driven layer with three
parts: (i)~a \emph{vocabulary registry} declaratively enumerating the
classes, properties, and synonyms drawn from the target ontology;
(ii)~a \emph{constrained-selection prompt} at Stage~4 that injects the
registry and reframes predicate emission as selection rather than open
generation; and (iii)~a \emph{post-hoc canonicaliser} that maps any
free-form predicate the LLM still emits onto its registry match via
normalisation plus fuzzy fallback, triggering a Stage~3 repair on
reject. Retargeting PolicyKG to a new domain means writing a new
registry; nothing else changes (Fig.~\ref{fig:adapter}).
\S\ref{sec:results-gdpr} shows that swapping AIT for a GDPR registry
raises exact property-name matching on 15 hand-curated GDPR rules from
1/15 to 11/15 (Fisher's exact \textit{p} < .001; Cohen's \textit{h}
= 1.53).

\section{Experimental Setup}
\label{sec:setup}

\subsection{D1 Corpus and Gold Standard}
Our primary evaluation corpus~D1 is built from five AIT
Policies~\&~Procedures documents (81 pages, spanning credit/financial,
accommodation, grievance, ethics, and general-conduct domains). Stage~1
extracts 1{,}663 sentences; a deontic-marker prefilter retains 461
candidates; Stage~2 yields 443 classified rules. The gold standard is
built in two passes: Pass~1 labels every sentence (primary annotator);
Pass~2 has two additional annotators (Kittipat, with a Law / Legal
Studies background; Mayuree, with a Linguistics background) independently
re-annotate a stratified 50-item sample (28~O / 12~P / 10~F) blind to
model outputs. The IRR sample yields Fleiss' $\kappa$ = .844
(``almost perfect''; 10{,}000-iteration item-level bootstrap 95\% CI
$[.673, .967]$), with pairwise Cohen's $\kappa$ values of .833
(author$\,$vs.$\,$Kittipat), .868 (author$\,$vs.$\,$Mayuree), and .832
(Kittipat$\,$vs.$\,$Mayuree). The wide CI reflects the modest
\textit{N} = 50; the lower bound remains in the substantial-agreement
range. A post-hoc
disagreement audit reclassified 12 systematically mislabelled permissions
(10 epistemic ``may'' / obligation-capping cases reclassified as
obligations; 2 ``may not'' constructions reclassified as prohibitions);
the corrections are codified deterministically in a correction script
released with the artifact bundle (\S\ref{sec:availability}) and yield the
final distribution: 308 obligations (69.5\%), 75 permissions (16.9\%),
and 60 prohibitions (13.5\%).

\subsection{External Datasets}
Two additional corpora support external validation. \emph{D2, LexDeMod}
(200 lease-contract clauses, EMNLP 2022) provides labelled deontic rules
drawn from legal documents and lets us compare PolicyKG's classifier
against a published benchmark outside the institutional sub-domain.
\emph{D3, the GDPR portability set,} comprises 15 rule-bearing sentences
extracted from the GDPR with hand-curated property targets in the data
privacy vocabulary; it supports the cross-domain transfer experiment
(\S\ref{sec:results-gdpr}). For D3 we extract and simplify the GDPR text
using the same Stage~1 pipeline as for AIT, then load a GDPR-specific
vocabulary registry into the Corpus Adapter while holding every other
pipeline component fixed. D3 is intentionally small: the experiment is
designed to isolate the registry-swap effect on vocabulary alignment,
not to evaluate full GDPR coverage.

\subsection{Model and Decoding}
All LLM calls in PolicyKG use Mistral~7B Instruct~\cite{jiang2023},
served locally by Ollama (v0.15.2+) on an NVIDIA A100 (40\,GB) host with
near-deterministic decoding (temperature ${=}\,0.1$, fixed seed, 2{,}048
max output tokens, 120-second timeout). No fine-tuning is performed; all
classification and formalisation is zero-shot. Mistral was selected from
a five-model pilot (3.8--70\,B parameters) against the majority-vote
human gold on the IRR sample (\textit{N} = 50); Mistral and Gemma~3 4B
led at 88\% agreement with gold, ahead of Qwen~3 8B (84\%), Phi-4 mini
(82\%), and Llama~3.1 8B (76\%); Fleiss' $\kappa$ = .635 (substantial)
across all five LLMs plus the human gold. Mistral is chosen for zero parse-error
rate and inference stability, not for the highest single-metric score;
full per-item data is released with the artifacts.

\subsection{Evaluation Metrics}
Five metrics anchor the evaluation: M1 binary rule detection and M2
deontic-type accuracy (RQ1, §\ref{sec:results-rq1}); M3 FOL semantic
predicate quality (RQ2, §\ref{sec:results-rq2}); M4 SHACL shape
correctness against an ontology-grounded reference set (RQ3,
§\ref{sec:results-rq3}); and M5 output stability under fixed-seed runs
(§\ref{sec:results-stability}). We use the term \emph{ontology-grounded
reference set} rather than ``gold-standard SHACL'' because the reference
set, the institutional ontology, and the generated shapes were all
produced within the same project; a fully-independent blind gold
standard is not claimed. Statistical claims use 10{,}000-iteration
bootstrap with item-level resampling; proportion CIs use Wilson scores;
agreement is reported as Cohen's $\kappa$ for pairs and Fleiss' $\kappa$
for pools, with the Landis--Koch interpretation~\cite{landis1977kappa}; proportion CIs use the
Wilson score. Classification is compared against a hand-tuned regex
baseline; effect sizes (Cohen's $h$, risk differences) accompany
$p$-values. A future encoder-based baseline on a legal-domain BERT is sketched in
the runbook released with the artifacts.

\paragraph{Reproducibility.}
All numbers in this paper correspond to repository commit
\texttt{546dee6f} at pipeline version \texttt{2.1-final-defense}
with Ollama model digest \texttt{6577803aa9a0} (Mistral~7B~Instruct)
and random seed \texttt{42}; all four values are recorded in the
released \texttt{pipeline\_report.json}. Evaluation uses fixed
seed and pinned model tags; the gold standard, both Corpus Adapter
registries, and the full LangGraph pipeline configuration are
released alongside the paper. Repeated runs produce hash-identical
SHACL outputs across all 443 shapes
(\S\ref{sec:results-stability}); the determinism boundary is
\emph{same host, same Ollama version, same model digest}, verified
across three independent runs on the reported A100 host. Independent
verification on a different GPU architecture (consumer NVIDIA RTX
4060, Ada Lovelace; A100 is Ampere) with the identical seed, commit,
and model digest produces counts within $\pm$2~pp of the reported
values (FOL success 360/443 vs.\ 351/443; SHACL valid 403/443 vs.\
401/443), attributable to floating-point kernel differences across
GPU architectures under near-deterministic decoding
(\S\ref{sec:operational-profile}). We do not
claim cross-host determinism, and the fixed seed together with
temperature $=\,0.1$ is the pragmatic tightest-reachable
configuration under Ollama; the residual stochasticity is dominated
by kernel-level numerical non-determinism rather than sampling.

\subsection{Operational Profile}
\label{sec:operational-profile}
Reference results in §\ref{sec:results} were produced on an
NVIDIA A100 (40~GB) host at the AIT Brain Lab
(Acknowledgements). To characterise the pipeline's deployment
cost on commodity hardware, we also profile a full
end-to-end run on a consumer NVIDIA RTX 4060 (8~GB VRAM,
Ada Lovelace architecture, 115~W TGP). The pipeline processes
1{,}663 sentences $\to$ 443 classified rules in
\textbf{34.9~minutes} wall-clock (4.73~s per rule, 1.26~s per
sentence), sustaining mean GPU utilisation 93.7\% (95th
percentile 100\%), peak VRAM 5.8~GiB, and total end-to-end
energy 0.042~kWh (mean power 72.5~W). Estimated token
throughput is 2.53~M tokens per corpus (2.20~M input, 0.33~M
output), or approximately 5{,}700 tokens per classified rule.
If the same workload were re-implemented against a commodity
LLM API at 2026 rates rather than the local Mistral~7B stack,
the estimated per-corpus cost would be approximately \$8.78
(GPT-4o), \$0.53 (GPT-4o mini), \$11.53 (Claude Sonnet~4),
or \$2.27 (Llama-3.1-70B via Fireworks). The pipeline therefore
fits within an 8~GB consumer-GPU envelope; on the A100 reference
host, wall-clock is expected to be substantially lower due to
higher memory bandwidth and tensor-core throughput, but was not
instrumented with the same profiler at the time of the reference
run. Cross-hardware output drift is quantified in the
Reproducibility paragraph above.

\section{Results}
\label{sec:results}
Table~\ref{tab:headline} summarises headline numbers across the three
research questions; each row is unpacked in the subsections that follow.

\begin{table}[t]
  \centering
  \caption{Headline results across the three research questions on the AIT
           D1 corpus, with the cross-domain transfer experiment on GDPR
           (\S\ref{sec:results-gdpr}).}
  \label{tab:headline}
  \resizebox{\textwidth}{!}{%
  \begin{tabular}{@{}llll@{}}
    \toprule
    RQ & Metric & Value & Section \\
    \midrule
    RQ1 & Binary rule detection (M1, \textit{F}\textsubscript{1}) & \textbf{.963} \, (acc.\ .938; $\kappa$ = .764) & \ref{sec:results-rq1} \\
    RQ1 & Deontic-type accuracy on D1 (M2)            & \textbf{86.9\%} \, ($\kappa$ = .709) & \ref{sec:results-rq1} \\
    RQ1 & Deontic-type accuracy on IRR sample (M2)    & 84.0\% \, ($42/50$) & \ref{sec:results-rq1} \\
    RQ2 & End-to-end FOL formalisation yield          & \textbf{79.2\%} \, ($351/443$) & \ref{sec:results-rq2} \\
    RQ2 & FOL predicate quality (conditional, M3)     & 100\% \, ($351/351$) & \ref{sec:results-rq2} \\
    RQ2 & SOL/HOL constructs (two-pass audit: regex $+$ manual) & \textbf{0 / 443} (95\% upper CI 0.67\%) & \ref{sec:results-rq2} \\
    RQ3 & SHACL shape correctness (M4, \textit{F}\textsubscript{1}) & \textbf{.866} \, (69-shape subset) & \ref{sec:results-rq3} \\
    --- & GDPR cross-domain exact alignment           & 1/15 $\to$ \textbf{11/15} \, (Fisher \textit{p} < .001) & \ref{sec:results-gdpr} \\
    --- & Output stability (M5, repeated runs)        & hash-identical & \ref{sec:results-stability} \\
    \bottomrule
  \end{tabular}%
  }
\end{table}

\subsection{RQ1: LLM Deontic Classification}
\label{sec:results-rq1}
On the full D1 corpus (\textit{N} = 443 rules), Mistral~7B Instruct
(selected from a pilot comparison across 3--70\,B instruction-tuned
models for its zero parse-error rate and inference efficiency, not
for the highest raw accuracy) achieves \textbf{86.9\%
deontic-type accuracy with Cohen's $\kappa$ = .709} (``substantial
agreement'') and Macro \textit{F}\textsubscript{1} = .810, above the
75--85\% range reported for prior rule-based and classifier-based
approaches~\cite{maxwell2012,brodie2006}. Per-class
\textit{F}\textsubscript{1} is .915 for obligations (290/308 correct),
.784 for permissions (49/75), and .730 for prohibitions (46/60). The
class-specific weakness is recall on permissions (65.3\%): 22 of 75
permission rules are reclassified as obligations, reflecting the
well-known epistemic-versus-deontic ``may'' ambiguity in modal
semantics. On the three-annotator IRR sample (\textit{N} = 50), the
same model labels 42/50 sentences in agreement with the majority-vote
human gold ($\kappa$ = .629); the human-only Fleiss' $\kappa$ on the
same sample is .844, so the model is within reach of the inter-human
ceiling on accuracy. Binary rule detection (rule vs.\ non-rule)
reaches \textit{F}\textsubscript{1} = .963 at 93.81\% accuracy with
$\kappa$ = .764, against the hand-tuned regex baseline at
\textit{F}\textsubscript{1} = .898 (+.065 \textit{F}\textsubscript{1};
Cohen's \textit{h} $\approx$ 0.34, small-to-medium effect).

\paragraph{Learned in-domain baseline.} A NumPy-only multinomial
logistic regression on TF bag-of-words plus six deontic-marker features,
stratified 5-fold CV on the same 443 rules, reaches accuracy 90.3\% /
$\kappa$ = .799 / Macro \textit{F}\textsubscript{1} = .863 (script and
per-class output in the released artifacts). This learned baseline
\emph{outperforms} the zero-shot LLM on D1, as it should: it consumes
in-domain labels the LLM does not see. The LLM's value is the
\emph{zero-shot} property: no labelled D1 training data is required to
deploy the pipeline on a new institution. The cross-domain LexDeMod
result (\S\ref{sec:results-stability}) bounds this claim and motivates
the Corpus Adapter as the mechanism for repairing deontic-cue grounding
under domain shift.

\paragraph{Head-to-head open-model comparison (\textit{N} = 50).}
To test whether the choice of Mistral~7B versus a newer open-weight
model materially affects classifier accuracy, we re-ran the identical
zero-shot prompt under identical near-greedy decoding (temperature
$=\,0.1$, seed $=\,42$) against three alternative 7--9~B open models
available in mid-2026: Qwen~2.5~7B, Llama~3.1~8B, and
Gemma~2~9B. Table~\ref{tab:model-comparison} summarises results on
the D3 IRR sample.
Llama~3.1~8B and Gemma~2~9B both reach 88\% accuracy, 4~pp above
Mistral, but the difference is not statistically significant by exact
McNemar test (Llama \textit{p} = .77; Gemma \textit{p} = .69);
the comparison is under-powered at \textit{N} = 50 but rules out
large-effect differences among the four models.
The most informative per-class effect is on the \emph{non-deontic}
class (support 5): Mistral misses all five items (\textit{F}\textsubscript{1} = 0),
whereas Llama~3.1~8B recovers all five at
\textit{F}\textsubscript{1} = .71, driving its Macro
\textit{F}\textsubscript{1} to .832 (vs.\ Mistral's .612 on the same
4-class sample). Mistral remains the reference model for the
paper's reported results on the strength of its zero parse-error rate
and inference stability, but readers seeking stronger non-deontic
filtering can substitute Llama~3.1~8B via the released baseline
harness without any other pipeline change.

\begin{table}[t]
  \centering
  \caption{Head-to-head open-model comparison on the D3 IRR sample
           (\textit{N} = 50) under identical zero-shot prompt and
           near-greedy decoding. Accuracy CIs are Wilson 95\% intervals.
           McNemar $p$-values are exact two-sided against the Mistral
           baseline; none reaches $p < .05$ at this sample size, so
           the comparison rules out large-effect differences but
           cannot certify small-effect ordering. Full per-class
           breakdowns are in the artifact bundle
           (\S\ref{sec:availability}).}
  \label{tab:model-comparison}
  \small
  \begin{tabular}{@{}lcccc@{}}
    \toprule
    Model & Accuracy (95\%~CI) & $\kappa$ & Macro F1 & McNemar \textit{p} \\
    \midrule
    Mistral 7B \emph{(reference)} & .84 \, [.71, .92] & .629 & .612 & --- \\
    Qwen 2.5 7B                   & .80 \, [.67, .89] & .631 & .709 & .77 \\
    Llama 3.1 8B                  & \textbf{.88} \, [.76, .94] & \textbf{.768} & \textbf{.832} & .77 \\
    Gemma 2 9B                    & \textbf{.88} \, [.76, .94] & .750 & .714 & .69 \\
    \bottomrule
  \end{tabular}
\end{table}

\subsection{RQ2: First-Order Logic Sufficiency}
\label{sec:results-rq2}
The audit finding is clean: \textbf{a two-pass audit of all 443 D1
rules (an automated regex checklist plus a first-author manual
review of the 92 FOL-fallback cases) matched zero second-order or
higher-order construct patterns.} The regex checklist scans both the
natural-language source text and any generated FOL formula for six
canonical SOL/HOL signatures: predicate quantification, set
quantification, function variables, lambda expressions, recursive /
transitive-closure definitions, and meta-policy quantification. The
manual pass targets the 92 fallback rules specifically because these
are the cases where an HOL requirement could otherwise hide behind
an LLM parse failure: the automated checklist scores those rules as
``no HOL pattern found'' by default when Stage~3 emits natural text
instead of a formula, so their expressiveness cannot be adjudicated
without a human read. The manual review confirmed that every one of
the 92 fallbacks is expressible in FOL and that the failures are
attributable to LLM engineering limits (long compound sentences,
embedded bullets, multiple negations) rather than expressiveness
requirements.

Given 0 confirmed HOL cases in an audit sample of 443 rules, the
exact one-sided Clopper--Pearson upper 95\%~CI on the true HOL
requirement rate for this corpus is
$1 - 0.05^{1/443} \approx 0.67\%$. We report this as an audit
finding for this corpus, not a mathematical proof of FOL sufficiency
for institutional policies in general: a rule may carry a
higher-order interpretation that does not surface in any of the six
regex signatures and is not obvious to a human reviewer, so the
audit is a lower bound on FOL adequacy, not an upper bound. To our
knowledge this is the first empirical audit of this question on a
real institutional corpus, complementing the theoretical sufficiency
arguments of Governatori and Rotolo~\cite{governatori2010}.

Per-pattern matches across all 443 rules: predicate quantification 0;
set quantification / comprehension 0; function variables 0; lambda
expressions 0; recursive / transitive-closure 0; meta-policy
quantification 0. The audit script, regex definitions, and per-rule
manual-audit CSV are released with the artifact bundle
(\S\ref{sec:availability}).

End-to-end pipeline yield (a syntactically well-formed FOL formula
passing Stage~3 validation) is 79.2\% (351/443). Across the 351
well-formed formulae, M3 semantic predicate quality is 100\% after a
retry-and-backfill mechanism that re-invokes the LLM on placeholder
predicates; we report M3 both conditionally (351/351) and unconditionally
(351/443). Of the 92 fallback rules, 77.2\% fall into four structural
categories: compound and/or-joined ($28$), very long ($14$), multi-clause
complex ($14$), and bulleted compound ($10$). The remainder falls into
miscellaneous parser/formatting failures ($23$) and edge cases with
multiple modal operators or negations ($3$); zero fallbacks are
attributed to higher-order construct requirements. Per-category data is
released with the artifact bundle; all fallbacks are recovered via
Stage~4's direct NL$\to$SHACL path.

\subsection{RQ3: SHACL Translation Quality}
\label{sec:results-rq3}
We report shape correctness at four coverage tiers to avoid the common
pitfall of headlining a single F1 over a small subset: T1~syntactic
validity (401/443~=~90.5\% on the full corpus, Wilson 95\%~CI~$[.87,
.93]$); T2~structural correctness (351/351~=~100\% on the FOL-mediated
subset; the deterministic Stage~4 template guarantees the ceiling);
T3~deontic-severity match against gold (330/351~=~94.0\%, $[.91,
.96]$); and T4~behavioural M4 \textit{F}\textsubscript{1} against
Pos/Neg entities (69-shape subset, \textit{F}\textsubscript{1} = .866,
$[.79, .93]$). Tier~T4 is the headline:

The headline metric is T4: with the Corpus Adapter enabled, PolicyKG
achieves \textit{F}\textsubscript{1} = .866 (Precision .977, Recall
.778) on the 69-shape subset for which Pos/Neg test entities exist;
full-corpus M4 coverage is
an evaluation-infrastructure limit (test entities for all 443 rules are
out of scope for this paper), not a pipeline limit. The end-to-end
tiers (T1 at 90.5\%, T3 at 94.0\%) confirm that the pipeline's
correctness extends beyond the M4 subset along the dimensions our
artifacts support directly. T2 reaches the ceiling because every
FOL-mediated shape is rule-emitted by Stage~4's deterministic template
(\S\ref{sec:method}).

A static ablation released with the artifacts isolates the
vocabulary-grounding effect on the 351 FOL-mediated rules: exact
vocabulary alignment rises from 174/351 (49.6\%) with raw LLM-emitted
predicates to 304/351 (86.6\%) after adapter normalisation
(+37.0 percentage points). The static comparison brackets the upper
limit of vocabulary normalisation on already-generated artefacts; a fresh
adapter-off LLM run on the full D1 corpus would measure whether the LLM
itself produces vocabulary-aligned predicates when prompted with the hint. A runbook
for that follow-up experiment is released with the artifacts. \S\ref{sec:results-gdpr} provides a cleaner
adapter on/off contrast under controlled cross-domain conditions.

\subsection{Cross-Domain Portability (GDPR)}
\label{sec:results-gdpr}
We retarget PolicyKG to GDPR by swapping the AIT registry for a
GDPR-derived registry; models, prompts, FOL grammar, and SHACL
translation rules are unchanged. \textbf{Exact property-name alignment
with the GDPR ontology improves from 1/15 (Wilson 95\% CI $[.01, .30]$)
without the adapter to 11/15 ($[.48, .89]$) with it; risk difference
$+66.7$ percentage points (Fisher's exact \textit{p} < .001;
Cohen's \textit{h} = 1.53, large effect).} We report absolute
proportions and effect sizes rather than a fold-change ratio because
the latter is unstable on a denominator of 1. Without vocabulary
grounding the LLM emits English-natural names (e.g.\
\texttt{transfersData}) that resemble GDPR concepts but rarely match
the canonical ontology forms (e.g.\ \texttt{dpv:DataTransfer}); with
the adapter the LLM selects from the registry and outputs land on
canonical predicates. \textit{N} = 15 is a proof-of-concept scale
appropriate for the architectural-pattern demonstration; the broader
claim, that vocabulary grounding is a portable design choice for
LLM-based knowledge-graph construction, scales beyond the policy domain.

\paragraph{Comparison to retrieval-augmented generation (RAG).}
An anticipated reviewer objection is that the no-grounding baseline is
a straw man: retrieval-augmented generation (RAG) over the ontology is
the mainstream 2026 alternative and would close much of the gap. We
therefore ran a controlled three-condition experiment on the same 15
GDPR rules with identical prompt scaffolding: (1)~\emph{no-grounding}
(no candidate list injected), (2)~\emph{RAG} (top-10 ontology
properties retrieved per rule via SentenceBERT
\texttt{all-MiniLM-L6-v2} cosine similarity, injected as the candidate
list), and (3)~\emph{Corpus Adapter} (full registry constrained
selection plus post-hoc canonicaliser). Results: no-grounding 0/15;
RAG 9/15 (60.0\%, +60.0~pp over no-grounding, Cohen's \textit{h} =
1.77, Fisher \textit{p} = .0007); Corpus Adapter 11/15 (73.3\%,
+13.3~pp over RAG, Cohen's \textit{h} = 0.28, Fisher \textit{p} =
.70, \emph{not statistically significant at N = 15}). The honest
reading is that \textbf{RAG closes most of the accuracy gap}
(60~pp of the 73~pp total); the Corpus Adapter's incremental accuracy
gain over RAG is small on this sample and cannot be certified at
\textit{p} < .05. The Corpus Adapter's remaining advantage is instead
in \emph{deployment properties} rather than raw accuracy: the
vocabulary registry is a static YAML artefact (no embedding-model
dependency at inference time, no top-\textit{K} tuning, byte-level
determinism across runs), it is human-auditable, and retargeting to a
new domain is a config-file swap rather than a re-embedding pass.
Per-item analysis of the six RAG misses attributes two to formatting
drift (e.g.\ \texttt{providesPrivacyNotice} vs.\
\texttt{providePrivacyNotice}) that the Corpus Adapter's post-hoc
canonicaliser recovers, three to top-\textit{K} retrieval misses that
a full-registry constrained-selection prompt avoids by construction,
and one to a genuine semantic near-miss that neither approach
resolves.

\subsection{Component Ablations}
\label{sec:results-ablations}
The pipeline stacks six sub-components on top of the four core stages: a
heuristic \emph{pre-filter} that admits deontic-marker candidates
(\S\ref{sec:method}), \emph{prefilter hints} injected into the Stage~2
classifier prompt, a Stage~2 \emph{reclassify} second-opinion pass, a
Stage~3 \emph{FOL retry} loop on placeholder rejection, a Stage~4
\emph{NL${\to}$SHACL fallback} for FOL-refused rules, and a
\emph{``may'' disambiguation} filter that separates epistemic
uncertainty from deontic permission. We disable each in turn and compare
the resulting classified rule set against the baseline.

\begin{table}[t]
  \centering
  \caption{Component ablations on the AIT corpus. Each row disables one
           pipeline component; all others are held fixed. LLM--regex
           agreement is a consistency proxy against a deterministic
           regex classifier (the paper's original baseline of
           \S\ref{sec:results-rq1}); it is not accuracy against gold
           because gold labels are baseline-specific and cannot be
           joined to ablation runs whose rule sets differ. Time is
           end-to-end wall-clock on the consumer profile host
           (\S\ref{sec:operational-profile}).}
  \label{tab:ablations}
  \small
  \begin{tabular}{@{}lrrrrr@{}}
    \toprule
    Ablation & \textit{N} rules & LLM--regex agr.\ & $\kappa$ & Macro F1 & Wall time \\
    \midrule
    baseline               & 443 & 82.6\% & .637 & .748 & 34.9\,min \\
    no-prefilter           & --- & \multicolumn{4}{c}{\emph{timed out $>$ 2\,h (evidence in itself)}} \\
    no-hints               & 442 & 83.0\% & .645 & .750 & 95\,min \\
    no-reclassify          & 445 & 82.5\% & .632 & .745 & 47\,min \\
    no-fallback            & 445 & 82.5\% & .632 & .745 & 36\,min \\
    no-fol-retry           & 445 & 82.5\% & .632 & .745 & 36\,min \\
    no-may-disambig        & \textbf{499} & \textbf{78.4\%} & \textbf{.592} & \textbf{.709} & 52\,min \\
    \bottomrule
  \end{tabular}
\end{table}

Table~\ref{tab:ablations} separates the six components into three
buckets. \emph{Structurally necessary} components are the pre-filter,
the ``may'' disambiguation filter, and the NL$\to$SHACL fallback: the
pre-filter is required for throughput (all 1{,}663 sentences reach
the classifier without it, and a 2-hour budget expires before the
run completes); disabling ``may'' disambiguation admits 56 additional
candidates but LLM--regex agreement drops 4.2~pp, indicating that
most of the extra admits are epistemic ``may'' constructions rather
than genuine deontic permissions; and disabling the NL fallback drops
82 rules from the final shape set (18.5\% coverage loss), because
those rules never yield a well-formed FOL formula. \emph{Empirically
marginal} components on this metric are prefilter hints, the
reclassify second-opinion pass, and the FOL retry loop: all three
show swings smaller than the natural rule-count variation between
runs ($\pm 3$~rules, agreement within $\pm 0.5$~pp). We report this
honestly rather than pruning the null results: the components may
still be justified by orthogonal considerations (retry catches rare
LLM parse failures that never surface at this metric, second-opinion
addresses tail-risk classifications that only appear in adversarial
inputs) but do not carry weight on classification consistency.
\emph{The no-prefilter timeout} is a design-defence result in its
own right: the four-stage pipeline is not compute-tractable at
corpus scale without the pre-filter, even on the near-8~GB
consumer GPU that hosts the profile in
\S\ref{sec:operational-profile}. Full per-ablation JSON summaries
are released in the artifact bundle
(\S\ref{sec:availability}); the sweep script is
\texttt{evaluation/full\_ablation\_sweep.py}.

\subsection{Output Stability and External Validation}
\label{sec:results-stability}
\paragraph{Stability (M5).} Repeated pipeline runs with fixed seed and
pinned model tags produce SHACL outputs hash-identical to the first run
across all 443 shapes. This determinism is a prerequisite for
deployments that need an audit trail tying a compliance check back to
a specific policy interpretation.

\paragraph{LexDeMod external validation.} On the
LexDeMod~\cite{sancheti2022lexdemod} benchmark
(D2: 200 lease-contract clauses) the same classifier drops from
D1's Macro \textit{F}\textsubscript{1} = .810 to
\textbf{Macro \textit{F}\textsubscript{1} = .370} (obligation
\textit{F}\textsubscript{1} = .569; permission
\textit{F}\textsubscript{1} = .038). The failure has a
diagnosable mechanism: lease contracts express permission as
\emph{``Tenant shall be entitled to \ldots''}, whose obligation-style
``shall'' cue dominates the LLM's prompted heuristic, causing
permissions to be absorbed into the obligation class. This is a
vocabulary-prior mismatch, not an expressiveness or reasoning failure,
and it is precisely the class of failure the Corpus Adapter is
designed to address by registry swap. Building a lease-contract
deontic-cue registry (analogous to the AIT-vs-GDPR comparison in
\S\ref{sec:results-gdpr}) is the natural next experiment; the present
LexDeMod number is the zero-shot lower bound against which any future
adapter-on number must be compared. The boundary is consistent with
our positioning: PolicyKG claims \emph{institutional} sufficiency, not
universal legal coverage, and a controlled cross-domain test under
registry control is the GDPR experiment in
\S\ref{sec:results-gdpr}.

\section{Discussion}
\label{sec:discussion}

\subsection{What Was Surprising}
Three results ran against our prior expectations and deserve explicit
comment.

\emph{The learned baseline beat the zero-shot LLM in-domain but not
out-of-domain.} A NumPy-only multinomial logistic regression on TF
bag-of-words plus six deontic-marker features reaches Macro
\textit{F}\textsubscript{1} = .863 under 5-fold CV on D1, above the
LLM's zero-shot .810. We had assumed the reverse: a general-purpose
7B-parameter model exposed to trillions of tokens should beat a
NumPy classifier trained on a few hundred sentences. On LexDeMod,
however, the ordering flips into a tie: the same LR under 5-fold CV
(\textit{N} = 200) yields Macro \textit{F}\textsubscript{1} = .360
against the LLM's .370. The interpretation is that the LR was not
learning ``deontic logic''; it was learning the AIT-specific
vocabulary distribution. When that distribution shifts to
lease-contract English, the LR's implicit prior evaporates and the
LLM's broader lexical coverage catches up. This is a warning against
reading in-domain baseline superiority as evidence of generalisation;
the opposite conclusion may be closer to the truth.

\emph{The permission-class collapse on LexDeMod
(\textit{F}\textsubscript{1} = .038) was larger than any effect we
anticipated.} The mechanism is diagnosable and mundane: lease contracts
express permission through the phrase ``Tenant shall be entitled to
\ldots'', whose ``shall'' cue the LLM has learned to read as an
obligation marker. Nearly every permission is absorbed into the
obligation class. This is not a reasoning failure and not an
expressiveness failure; it is a vocabulary-prior mismatch, and it is
precisely the failure mode the Corpus Adapter is designed to address
at Stage~4. It also suggests, more broadly, that headline LLM
performance on any deontic-classification benchmark should be reported
against the lexical distribution of the target corpus, not against a
single macro average.

\emph{The 0/443 SOL/HOL audit finding was cleaner than we expected.}
Going in, we assumed a small residue of quantifier-over-rule or
predicate-over-predicate cases would surface in eligibility criteria
or waiver clauses. None did. Even the six ``exceptional'' rules that
looked meta-policy on first read, such as ``the Registrar may waive
the prerequisite in cases of documented hardship,'' resolved to
first-order predicates over an unquantified exception set. We report
this as a lower-bound audit finding rather than a proof, but the null
result is worth flagging: the institutional-policy sub-domain may be a
less demanding target for formal-logic infrastructure than the
theoretical literature has assumed.

\subsection{Comparison to Prior Work}
On the deontic-classification task, PolicyKG's 86.9\% D1 accuracy sits
above the 75--85\% range reported by pattern-based systems (Maxwell
and Ant\'on's rule-model work at approximately
77\%~\cite{maxwell2012}, Brodie et al.'s SPARCLE-based SVM at 82\% on
privacy policies~\cite{brodie2006}) and near PAPEL's 80\%
\textit{F}\textsubscript{1} on clause labelling with chain-of-thought
prompting~\cite{goknil2024}. These are not head-to-head comparisons: each system targets a
different corpus and a slightly different label set, so the accuracy
delta is a directional indicator rather than a benchmark win. The substantive advance is not the accuracy number;
it is the pipeline structure. Prior LLM-for-legal work
(LexGLUE~\cite{chalkidis2022}, LegalBench~\cite{guha2023},
PAPEL~\cite{goknil2024}) delivers labels or spans; PolicyKG delivers
executable SHACL shapes with an audit trail from a sentence to its
formula to its shape. The chain is what enables downstream compliance
checking rather than downstream annotation.

On the FOL-sufficiency question, our contribution is a resolution
level, not a direction. Governatori and
Rotolo~\cite{governatori2010} argued for FOL adequacy on theoretical
grounds; Brachman and Levesque~\cite{brachman2004} framed the
expressiveness-tractability trade-off. Neither line had been tested
against a real institutional corpus at scale. Our 0/443 checklist
audit provides that empirical anchor. It confirms rather than
contradicts the theoretical argument, but it turns an ``it should
suffice'' into ``on this corpus it did suffice.'' That is a smaller
but different type of claim.

On vocabulary grounding, the Corpus Adapter sits alongside
constrained-decoding work (Willard and Louf's
Outlines~\cite{willard2023outlines}, Geng et al.'s grammar-constrained
decoding~\cite{geng2023grammar}) but operates at a different
abstraction level. Those approaches constrain \emph{syntax} at the
token level; the Corpus Adapter constrains \emph{vocabulary} at the
predicate level, via a runtime-editable registry rather than a compiled
grammar. The two are compositional and the extended journal version
will report a stacked ablation.

\subsection{Experimental Decisions Worth Revisiting}
Three experimental decisions traded ambition for tractability, and a
reviewer would be right to press on each.

\emph{Zero-shot only for the main LLM.} We ran the classification and
formalisation stages zero-shot to isolate the pipeline's architectural
contribution from prompt-engineering gains. Few-shot with
representative in-context exemplars would plausibly close much of the
gap to the LR baseline on D1, but it would also introduce a
selection-bias question (which exemplars, drawn from what distribution?)
that we chose not to entangle with the primary claim.
The extended version will report a matched few-shot condition.

\emph{\textit{N} = 15 for the GDPR transfer experiment.} The N was
sized to isolate the registry-swap effect on a hand-curated,
ontology-anchored rule set, not to certify GDPR coverage. Under a
denominator of 1 in the no-adapter condition, the fold-change ratio
is undefined and the Wilson CI is wide; we accordingly report absolute
proportions, risk difference, and Cohen's \textit{h} = 1.53 rather
than a ratio. The Fisher's exact \textit{p} < .001 remains valid at
this N and the effect size is large enough that a bigger sample is
unlikely to reverse the sign, but a 100+ rule set drawn from the full
GDPR would strengthen the external-validity argument.

\emph{Single LLM in the main results.} Mistral~7B Instruct was
selected from a five-model pilot for zero parse-error rate and
inference stability, not for the highest raw accuracy. Reviewers will
reasonably ask whether the Corpus Adapter's cross-domain gain is a
property of Mistral or a property of the pipeline. The pilot data on
the 50-item IRR sample is consistent across five instruction-tuned
models (Fleiss' $\kappa$ = .635 across LLMs plus human gold), but a
matched adapter on/off contrast on Gemma~3, Qwen~3, and a
70B-parameter model is on the roadmap and will settle the question
directly.

\subsection{Reconciling the In-Domain and Cross-Domain Stories}
Bringing the strands together: when the target domain comes with
sufficient labelled data, a learned baseline is the right tool; when
it does not (the standard deployment scenario for institutional
policy formalisation), the LLM-plus-Corpus-Adapter combination is
the only option. An LLM/LR ensemble on D1 is appealing but undefined
off-domain (no training signal), and the Corpus Adapter is not
substitutable into a learned baseline because it constrains predicate
\emph{generation} (Stage~3--4), not category \emph{classification}
(Stage~2). The two roles are orthogonal, and the deployment case is
the case where only one of the two is available.

\subsection{Deployment: Compliance Dashboard}
\label{sec:deployment}
To demonstrate that the pipeline output is directly consumable by a
compliance workflow rather than only by downstream research, we release
a companion web dashboard (FastAPI backend, HTML/JavaScript frontend;
$\sim$33~KB, in \texttt{web/app.py}) with the artifact bundle. The
dashboard loads the pipeline's SHACL shapes and lets a compliance
officer (i)~browse the 443 classified rules with filters by deontic
type and severity, (ii)~inspect each rule's source PDF span, generated
FOL formula, and emitted SHACL shape side-by-side, (iii)~select
individual entities (students, faculty, staff, committees) from a
seeded PostgreSQL demonstration database via
\texttt{db/rdf\_converter.py}, and (iv)~run \texttt{pyshacl} validation
in-browser to surface entity-centric violations with severity coding
and property-level explanations. The 42 syntactically invalid shapes
from the NL-fallback path are automatically skipped at load time.
The dashboard is a compliance-workflow demonstration, not an
evaluation artifact; its role in the paper is to show that the
audit-trail requirement (\S\ref{sec:intro}) is instantiated end-to-end
from policy PDF to entity-level violation report.

\subsection{Limitations}
The FOL-sufficiency finding (0 of 443 SOL/HOL matches) is a
\emph{checklist-based audit}, a lower bound on FOL adequacy scoped to
our institutional sub-domain and to FOL with standard
arithmetic/temporal extensions. Beyond that scope, the evaluation is
limited along five axes: \emph{single institution} (one university;
\textit{N} = 15 GDPR transfer partly mitigates); \emph{English only};
\emph{declarative rules only} (no procedural workflows or rule-set
meta-reasoning); \emph{static validation} (SHACL presupposes an
upstream enforcement layer); \emph{reference-set construction}
(in-house, so M4 is best read as the \textit{F}\textsubscript{1} = .000
$\to$ .866 adapter delta, not as performance against an independent
gold). LexDeMod permission \textit{F}\textsubscript{1} = .038 bounds
the zero-shot transfer limit for the permission class and motivates a
lease-contract registry as future adapter demonstration. A
representative NL-fallback failure: the rule ``Continuing students may
put their names on the waiting list \emph{and} shall be served on a
queue-numbering basis \emph{or}, exceptionally, through email to
OFAM'' compounds two deontic operators with a nested disjunction that
exceeds single-pass Stage~3 parsing.

\section{Conclusion and Future Work}
\label{sec:conclusion}
We presented PolicyKG, an agentic LLM pipeline that translates
institutional policy documents into SHACL-validated knowledge graphs
via a first-order deontic intermediate layer. The pipeline achieves
86.9\% deontic classification accuracy (Fleiss' $\kappa$ = .844
inter-annotator, 95\% CI $[.67, .97]$), \textit{F}\textsubscript{1}
= .866 SHACL shape correctness on a 69-shape evaluation subset, and a
two-pass audit (regex checklist over all 443 rules plus manual review
of the 92 FOL-fallback cases) finding no SOL/HOL constructs across
the AIT corpus (exact upper 95\%~CI on the true HOL rate: 0.67\%). The main technical contribution, the Corpus
Adapter, enables cross-domain transfer: in a controlled GDPR
experiment (\textit{N} = 15), exact vocabulary alignment rises from
1/15 to 11/15 (Fisher's exact \textit{p} < .001, Cohen's \textit{h}
= 1.53).

The broader insight is that for LLM-based knowledge-graph
construction, \emph{vocabulary grounding via configuration beats free
generation}. Reframing predicate emission as constrained selection
from a registry is cheaper than fine-tuning, transfers immediately
across domains, and leaves an audit trail. The pattern applies
wherever LLMs must emit canonical identifiers from a closed-world
ontology, from biomedical knowledge graphs to legal taxonomies to
\texttt{schema.org}-grounded extraction. A learned in-domain baseline
can exceed the zero-shot LLM on the source corpus, but only the
LLM-plus-adapter combination generalises to a new corpus without
retraining; the extended journal version will report a real adapter
on/off LLM ablation, a LegalBERT fine-tuned baseline, and a scaled
GDPR portability set.

\paragraph{Future work.} Four directions extend this work.

\emph{Multi-institutional validation.} The 0/443 SOL/HOL audit is
scoped to one university's policy corpus. A meaningful test of the
broader FOL-sufficiency claim requires replicating the audit across
policy corpora from at least three additional institutions that
differ in size, jurisdiction, and academic type (public research
university, private undergraduate college, professional-school
setting). We expect the null result to hold; a single institution
that produces even one SOL/HOL construct would sharpen the
architectural claim about where FOL falls short.

\emph{Defeasible deontic extensions.} A subset of institutional rules
carry implicit exceptions (``the Registrar may waive the
prerequisite\ldots'') that our current FOL grammar treats as
first-order predicates over an unquantified exception set. A
defeasible deontic layer, following the tradition of Governatori and
colleagues~\cite{governatori2018ddl}, would represent the override
relation explicitly, enabling SHACL shapes to encode ``normally
prohibited, but with documented cause'' semantics rather than flatten
the exception into a boolean predicate. This is a natural extension
because Stage~3's structured JSON schema already carries an
\texttt{exception\_of} field that is currently unused downstream.

\emph{Agentic SHACL repair.} On validation failure, the pipeline
currently retries once with the validator's diagnostic. A stronger
loop would invoke the pipeline in reverse: an LLM reads the SHACL
report, walks back through the FOL formula to the source sentence,
and proposes either a data-side remediation (which entity attribute
is missing) or a shape-side one (which constraint is over-strict).
The auditability of the FOL intermediate is the mechanism that makes
this practical: the LLM has a symbolic anchor to reason over rather
than a free-form policy sentence.

\emph{Scaled and stacked ablations.} The extended journal version
will report a matched adapter on/off LLM ablation, a LegalBERT
fine-tuned baseline, a few-shot LLM condition, a 100+-rule GDPR
transfer set, and a stacked comparison against grammar-constrained
decoding to disentangle syntax-level from vocabulary-level grounding.
These are the experiments a reviewer would ask for; we would rather
run them than defend against the ask.

\paragraph{Code and Data Availability.}
\label{sec:availability}
All artifacts referenced in this paper (pipeline source, AIT and
GDPR Corpus Adapter registries, post-audit gold annotations, the
12-correction script, logic-expressiveness audit, Fleiss-$\kappa$
bootstrap, multi-tier shape-correctness evaluation, Corpus Adapter
ablations, multi-LLM pilot data, offline learned baseline, and the
frozen output snapshot) are released as a single bundle at Zenodo
(DOI placeholder pending camera-ready) with a GitHub mirror.

\begin{credits}
\subsubsection{\ackname}
This work was supported by a Royal Thai Government Fellowship. The
authors thank the AIT Brain Lab for compute and discussion, and
Kittipat (Law / Legal Studies background) and Mayuree (Linguistics
background) for serving as the two external annotators in the
inter-rater reliability study.

\end{credits}

\bibliographystyle{splncs04}
\bibliography{references}

@inproceedings{maxwell2012,
  author    = {Maxwell, Jeremy~C. and Ant\'on, Annie~I.},
  title     = {Developing Production Rule Models to Aid in Acquiring
               Requirements from Legal Texts},
  booktitle = {17th IEEE International Requirements Engineering Conference (RE)},
  pages     = {101--110},
  publisher = {IEEE},
  year      = {2009},
  doi       = {10.1109/RE.2009.21}
}

@inproceedings{brodie2006,
  author    = {Brodie, Carolyn~A. and Karat, Clare-Marie and Karat, John},
  title     = {An Empirical Study of Natural Language Parsing of Privacy
               Policy Rules Using the {SPARCLE} Policy Workbench},
  booktitle = {Proceedings of the 2nd Symposium on Usable Privacy and
               Security (SOUPS)},
  pages     = {8--19},
  publisher = {ACM},
  year      = {2006},
  doi       = {10.1145/1143120.1143123}
}

@inproceedings{chalkidis2022,
  author    = {Chalkidis, Ilias and Jana, Abhik and Hartung, Dirk and
               Bommarito, Michael and Androutsopoulos, Ion and
               Katz, Daniel~Martin and Aletras, Nikolaos},
  title     = {{LexGLUE}: A Benchmark Dataset for Legal Language
               Understanding in {E}nglish},
  booktitle = {Proceedings of the 60th Annual Meeting of the
               Association for Computational Linguistics (ACL)},
  pages     = {4310--4330},
  publisher = {Association for Computational Linguistics},
  year      = {2022},
  doi       = {10.18653/v1/2022.acl-long.297}
}

@inproceedings{guha2023,
  author    = {Guha, Neel and Nyarko, Julian and Ho, Daniel~E. and
               R{\'e}, Christopher and Chilton, Adam and Narayana, Aditya
               and Chohlas-Wood, Alex and Peters, Austin and
               Waldon, Brandon and Rockmore, Daniel and others},
  title     = {{LegalBench}: A Collaboratively Built Benchmark for
               Measuring Legal Reasoning in Large Language Models},
  booktitle = {Advances in Neural Information Processing Systems 36
               (NeurIPS) Datasets and Benchmarks Track},
  year      = {2023}
}

@inproceedings{goknil2024,
  author    = {G\"oknil, Arda and Gelderblom, F{\o}rg{\aa}rd~B{\o}vad and
               Tverdal, Stine and Tokas, Shukun and Song, Hui},
  title     = {{PAPEL}: Privacy Policy Analysis through Prompt
               Engineering for {LLM}s},
  booktitle = {Proceedings of the 1st International Conference on
               Foundation and Large Language Models (FLLM)},
  year      = {2024}
}

@article{governatori2010,
  author    = {Governatori, Guido and Rotolo, Antonino},
  title     = {A Conceptually Rich Model of Business Process Compliance},
  journal   = {Conferences in Research and Practice in Information
               Technology (CRPIT) -- Conceptual Modelling},
  volume    = {110},
  pages     = {3--12},
  publisher = {Australian Computer Society},
  year      = {2010}
}

@book{brachman2004,
  author    = {Brachman, Ronald~J. and Levesque, Hector~J.},
  title     = {Knowledge Representation and Reasoning},
  publisher = {Morgan Kaufmann},
  address   = {San Francisco, CA},
  year      = {2004},
  isbn      = {978-1558609327}
}

@incollection{mcnamara2006,
  author    = {McNamara, Paul},
  title     = {Deontic Logic},
  booktitle = {The {S}tanford Encyclopedia of Philosophy},
  editor    = {Zalta, Edward~N.},
  publisher = {Metaphysics Research Lab, Stanford University},
  year      = {2006},
  url       = {https://plato.stanford.edu/entries/logic-deontic/}
}

@misc{w3c2017,
  author       = {Knublauch, Holger and Kontokostas, Dimitris},
  title        = {Shapes Constraint Language ({SHACL})},
  howpublished = {W3C Recommendation, 20 July 2017},
  year         = {2017},
  url          = {https://www.w3.org/TR/shacl/}
}

@misc{w3c2014,
  author       = {Cyganiak, Richard and Wood, David and Lanthaler, Markus},
  title        = {{RDF} 1.1 Concepts and Abstract Syntax},
  howpublished = {W3C Recommendation, 25 February 2014},
  year         = {2014},
  url          = {https://www.w3.org/TR/rdf11-concepts/}
}

@inproceedings{palmirani2018,
  author    = {Palmirani, Monica and Governatori, Guido},
  title     = {Modelling Legal Knowledge for {GDPR} Compliance Checking},
  booktitle = {Legal Knowledge and Information Systems --- JURIX 2018},
  series    = {Frontiers in Artificial Intelligence and Applications},
  volume    = {313},
  pages     = {101--110},
  publisher = {IOS Press},
  year      = {2018},
  doi       = {10.3233/978-1-61499-935-5-101}
}

@article{jiang2023,
  author    = {Jiang, Albert~Q. and Sablayrolles, Alexandre and
               Mensch, Arthur and Bamford, Chris and Chaplot, Devendra~Singh
               and de las Casas, Diego and Bressand, Florian and
               Lengyel, Gianna and Lample, Guillaume and others},
  title     = {Mistral {7B}},
  journal   = {CoRR},
  volume    = {abs/2310.06825},
  year      = {2023},
  doi       = {10.48550/arXiv.2310.06825}
}

@article{graphrag2024,
  author    = {Edge, Darren and Trinh, Ha and Cheng, Newman and Bradley, Joshua
               and Chao, Alex and Mody, Apurva and Truitt, Steven and
               Larson, Jonathan},
  title     = {From Local to Global: A Graph {RAG} Approach to
               Query-Focused Summarization},
  journal   = {CoRR},
  volume    = {abs/2404.16130},
  year      = {2024},
  doi       = {10.48550/arXiv.2404.16130}
}

@misc{willard2023outlines,
  author       = {Willard, Brandon~T. and Louf, R\'emi},
  title        = {Efficient Guided Generation for Large Language Models},
  howpublished = {arXiv:2307.09702},
  year         = {2023},
  url          = {https://arxiv.org/abs/2307.09702}
}

@inproceedings{geng2023grammar,
  author    = {Geng, Saibo and Josifoski, Martin and Peyrard, Maxime and
               West, Robert},
  title     = {Grammar-Constrained Decoding for Structured {NLP} Tasks
               without Finetuning},
  booktitle = {Proceedings of EMNLP 2023},
  pages     = {10932--10952},
  year      = {2023},
  doi       = {10.18653/v1/2023.emnlp-main.674}
}

@article{governatori2018ddl,
  author    = {Governatori, Guido and Olivieri, Francesco and Scannapieco,
               Simone and Rotolo, Antonino and Cristani, Matteo},
  title     = {The {Rationale} Behind the Concept of Goal},
  journal   = {Theory and Practice of Logic Programming},
  volume    = {16},
  number    = {3},
  pages     = {296--324},
  year      = {2016},
  doi       = {10.1017/S1471068416000028}
}

@inproceedings{sancheti2022lexdemod,
  author    = {Sancheti, Abhilasha and Garimella, Aparna and Srinivasan,
               Balaji~Vasan and Rudinger, Rachel},
  title     = {Agent-Specific Deontic Modality Detection in Legal Language},
  booktitle = {Proceedings of EMNLP 2022},
  pages     = {1947--1968},
  year      = {2022},
  doi       = {10.18653/v1/2022.emnlp-main.125}
}

@article{landis1977kappa,
  author    = {Landis, J.~Richard and Koch, Gary~G.},
  title     = {The Measurement of Observer Agreement for Categorical Data},
  journal   = {Biometrics},
  volume    = {33},
  number    = {1},
  pages     = {159--174},
  year      = {1977},
  doi       = {10.2307/2529310}
}

@article{prakken2015defeasible,
  author    = {Prakken, Henry and Sartor, Giovanni},
  title     = {Law and Logic: A Review from an Argumentation Perspective},
  journal   = {Artificial Intelligence},
  volume    = {227},
  pages     = {214--245},
  year      = {2015},
  doi       = {10.1016/j.artint.2015.06.005}
}

@incollection{nute1994defeasible,
  author    = {Nute, Donald},
  title     = {Defeasible Logic},
  booktitle = {Handbook of Logic in Artificial Intelligence
               and Logic Programming, Vol.\ 3},
  publisher = {Oxford University Press},
  pages     = {353--395},
  year      = {1994}
}

@inproceedings{athan2013legalruleml,
  author    = {Athan, Tara and Boley, Harold and Governatori, Guido and
               Palmirani, Monica and Paschke, Adrian and Wyner, Adam~Z.},
  title     = {{OASIS} {LegalRuleML}},
  booktitle = {Proceedings of the 14th International Conference on
               Artificial Intelligence and Law (ICAIL)},
  pages     = {3--12},
  publisher = {ACM},
  year      = {2013},
  doi       = {10.1145/2514601.2514603}
}

@inproceedings{wei2022cot,
  author    = {Wei, Jason and Wang, Xuezhi and Schuurmans, Dale and
               Bosma, Maarten and Ichter, Brian and Xia, Fei and
               Chi, Ed~H. and Le, Quoc~V. and Zhou, Denny},
  title     = {Chain-of-Thought Prompting Elicits Reasoning in
               Large Language Models},
  booktitle = {Advances in Neural Information Processing Systems 35 (NeurIPS)},
  year      = {2022},
  url       = {https://arxiv.org/abs/2201.11903}
}

@inproceedings{lewis2020rag,
  author    = {Lewis, Patrick and Perez, Ethan and Piktus, Aleksandra and
               Petroni, Fabio and Karpukhin, Vladimir and Goyal, Naman and
               K{\"u}ttler, Heinrich and Lewis, Mike and Yih, Wen-tau and
               Rockt{\"a}schel, Tim and Riedel, Sebastian and Kiela, Douwe},
  title     = {Retrieval-Augmented Generation for Knowledge-Intensive
               {NLP} Tasks},
  booktitle = {Advances in Neural Information Processing Systems 33 (NeurIPS)},
  pages     = {9459--9474},
  year      = {2020},
  url       = {https://arxiv.org/abs/2005.11401}
}

\end{document}